\pdfoutput=1

\documentclass[11pt]{article}

\usepackage[]{EACL2023}

\usepackage{times}
\usepackage{latexsym}

\usepackage[T1]{fontenc}

\usepackage[utf8]{inputenc}

\usepackage{microtype}

\usepackage{inconsolata}

\usepackage{url}
\usepackage[linesnumbered,ruled]{algorithm2e}
\usepackage{algorithmic}
\usepackage{multirow}
\usepackage{mathtools}

\title{PEACH: Pre-Training Sequence-to-Sequence Multilingual Models for Translation with Semi-Supervised Pseudo-Parallel Document Generation}

\newcommand*\samethanks[1][\value{footnote}]{\footnotemark[#1]}

\author{
  Alireza Salemi\textsuperscript{1}\thanks{equal contribution}, Amirhossein Abaskohi\textsuperscript{1}\samethanks, Sara Tavakoli\textsuperscript{1}, \\
  \textbf{Yadollah Yaghoobzadeh\textsuperscript{1},
  Azadeh Shakery\textsuperscript{1,2}} \\
  \textsuperscript{1}School of Electrical and Computer Engineering \\ College of Engineering, University of Tehran, Tehran, Iran \\
  \textsuperscript{2}School of Computer Science \\
  Institute for Research in Fundamental Sciences (IPM), Iran \\
  \small{\texttt{\{alireza.salemi,amir.abaskohi,saratavakoli77,y.yaghoobzadeh,shakery\}@ut.ac.ir}}
}

\begin{document}
\maketitle
\begin{abstract}
Multilingual pre-training significantly improves many multilingual NLP tasks, including machine translation. Most existing methods are based on some variants of masked language modeling and text-denoising objectives on monolingual data. Multilingual pre-training on monolingual data ignores the availability of parallel data in many language pairs. Also, some other works integrate the available human-generated parallel translation data in their pre-training. This kind of parallel data is definitely helpful, but it is limited even in high-resource language pairs. This paper introduces a novel semi-supervised method, SPDG, that generates high-quality pseudo-parallel data for multilingual pre-training. First, a denoising model is pre-trained on monolingual data to reorder, add, remove, and substitute words, enhancing the pre-training documents' quality. Then, we generate different pseudo-translations for each pre-training document using dictionaries for word-by-word translation and applying the pre-trained denoising model. The resulting pseudo-parallel data is then used to pre-train our multilingual sequence-to-sequence model, PEACH. Our experiments show that PEACH outperforms existing approaches used in training mT5 \citep{mt5} and mBART \citep{mbart} on various translation tasks, including supervised, zero- and few-shot scenarios. Moreover, PEACH's ability to transfer knowledge between similar languages makes it particularly useful for low-resource languages. Our results demonstrate that with high-quality dictionaries for generating accurate pseudo-parallel, PEACH can be valuable for low-resource languages.
\end{abstract}

\section{Introduction}



Machine Translation (MT) involves transferring a text from one language to another. Recent investigations have revealed that multilingual pre-training on a large corpus is profitable for NLP systems' performance on multilingual downstream tasks \citep{mbart, Lample2019CrosslingualLM, cc100-2, mt5, bert} and knowledge transferability between languages \citep{wu-dredze-2019-beto, KWMR20, mbart}. Furthermore, using parallel data in pre-training encoder and encoder-decoder models effectively increases the models' performance in downstream tasks \citep{Lample2019CrosslingualLM, chi-etal-2021-mt6}. The existing pre-training approaches are mainly based on Masked Language Modeling (MLM) and its variations \citep{mbart, t5, mt5, bart}. 

Although using parallel data in pre-training multilingual models improves their performance on downstream tasks, the amount of available parallel data is limited \citep{10.5555/3495724.3495910}. Moreover, MLM-based objectives for sequence-to-sequence (seq2seq) models usually ask the model to generate an output in the same language as input, which is not in the interests of translation tasks. Additionally, MLM-based objectives use shared subwords or alphabets between different languages to learn shared embedding spaces across them \citep{Lample2019CrosslingualLM, lample2017unsupervised,https://doi.org/10.48550/arxiv.1702.03859}; this would not be possible for languages without shared alphabets. 

Using dictionaries to define anchor points between different languages in cross-lingual pre-training of the encoder of seq2seq models has been investigated and shown to be effective for unsupervised translation \citep{duan-etal-2020-bilingual}. Still, it never has been used as a method for pre-training multilingual seq2seq models. 


Our proposed method, Semi-Supervised Pseudo-Parallel Document Generation (SPDG), addresses the challenge of limited parallel data for low-resource languages by leveraging dictionaries to generate pseudo-parallel documents. SPDG adopts unsupervised translation techniques \citep{kim-etal-2018-improving, lample2017unsupervised} to generate a high-quality translation for each pre-training document. We use a pre-trained denoising seq2seq model with word reordering, adding, removing, and substituting to enhance the quality of the word-by-word translated document. The improved unsupervised translated text is used as the target text for training our multilingual seq2seq model, PEACH, using SPDG as a new pre-training method. SPDG enables transfer of knowledge between similar languages, making it particularly useful for low-resource languages.

Our experiments show that PEACH outperforms the pre-trained models with mT5's MLM and mBART's MLM with Reordering objectives in English, French, and German. Additionally, PEACH demonstrates strong performance in zero- and few-shot scenarios. Moreover, we test our model for other multilingual tasks, such as natural language inference, to investigate the model's ability in this task. Our results show that our model achieves a higher score in this task than other objectives, which shows PEACH's ability to transfer knowledge between languages. The main contribution of this paper is twofold:
\begin{itemize}
    \item We propose a novel semi-supervised pre-training method using bilingual dictionaries and pre-trained denoising models for seq2seq multilingual models.
    \item We show the benefits of SPDG objective in translation, supervised and zero- and few-shot cases, and knowledge transfer between languages.
\end{itemize}

\section{Related Work}

Among the first endeavor for MT,  dictionary and rule-based methods were popular \citep{Dolan1993CombiningDA,kaji1988efficient,meyers-etal-1998-multilingual}, followed by Knowledge-Based Machine Translation (KBMT) and statistical methods \citep{mitamara1993automated, Carbonell1981StepsTK, koehn2009statistical, al1999statistical}. The popularity of neural machine translation has only grown in the recent decade with the introduction of the first deep neural model for translation \citep{kalchbrenner2013recurrent}.

While the RNN-based seq2seq models seemed to be promising in neural machine translation \citep{Wu2016GooglesNM,Bahdanau2015NeuralMT,10.5555/2969033.2969173}, the advent of the transformer architecture \citep{transformer} plays an integral role in modern MT. With the introduction of the transformer architecture, pre-training general-purpose language models seemed to be an effective way to improve different NLP tasks \citep{bert,roberta}. In most cases, transformer models were asked to denoise a noisy input to learn a language \citep{bart,bert,t5}. One of the most popular pre-training objectives for both encoder-only and encoder-decoder models is called Masked Language Modeling (MLM), in which the model should predict the masked part of a document and generate it in its output \citep{t5}. However, many other objectives were also developed for encoder-decoder and encoder-only models \cite{mass,electra}.

Meanwhile, unsupervised methods for neural machine translation (NMT) using monolingual corpora based on adversarial learning \citep{lample2017unsupervised} and transformer-based text denoising \citep{kim-etal-2018-improving} was tested and demonstrated promising outcomes. Using bilingual dictionaries for defining anchors in pre-training unsupervised translation models was successful \citep{duan-etal-2020-bilingual} but never has been used for generating data for supervised translation on a large scale. Our work differs from using dictionaries as anchor points for learning a better representation for tokens in encoder \citep{duan-etal-2020-bilingual}. We use dictionaries to generate a pseudo translation of the source language in the target language instead of just defining some anchor points. Thus, the model in pre-training steps learns to generate a text in the target language based on input in the source language using only monolingual data and dictionaries on a large scale.

Pre-training task-specific models by generating pseudo-summaries was successful in some cases for summarization, question answering, and speech recognition \citep{chen-etal-2017-reading,arman,pegasus,abaskohi2022automatic}, but it has not been performed for pre-training encoder-decoder seq2seq models for supervised translation according to the best of our knowledge. On the other hand, the endeavors for pre-training specific models for translation ended up in training multilingual language models \citep{mt5,mbart}. mT5 \citep{mt5} is trained with the MLM objective of T5 \citep{t5}. In its pre-training objective, some spans of the input document are masked by specific tokens, and the model has to predict those spans by generating them in its output. mBART \citep{mbart} is another multilingual model based on the BART \citep{bart} model, pre-trained with MLM with Reordering objective. In mBART's pre-training objective, the order of sentences in the input document is corrupted while a specific token masks some spans of the document. The model has to generate the original document in its output.

PEACH is different from both mentioned models because we use a semi-supervised method to generate several pseudo-translations (one for each selected  language) of each pre-training document. These translations are then fed to pre-train PEACH. Furthermore, in the mentioned models, the inputs and outputs are from the same language while we ask the model to translate texts from one language to another in our pre-training phase.

\begin{figure*}[!ht]
    \centering
    \includegraphics[scale=0.35]{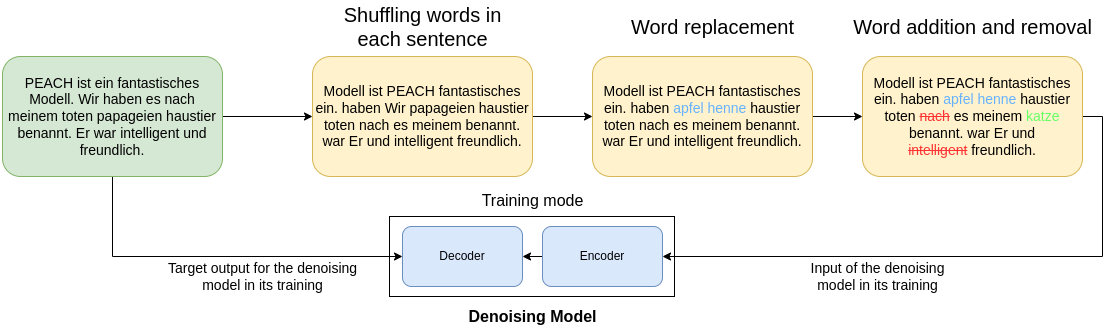}
    \caption{An overview of denoising objectives used for training denoising models. We use word shuffling, addition, substitution, and removing based on the values in Table \ref{appendix:table:denosing-rate} in Appendix \ref{appendix:denoising-details}.}
    \label{fig:denoising}
\end{figure*}

\begin{figure*}[!ht]
    \centering
    \includegraphics[scale=0.35]{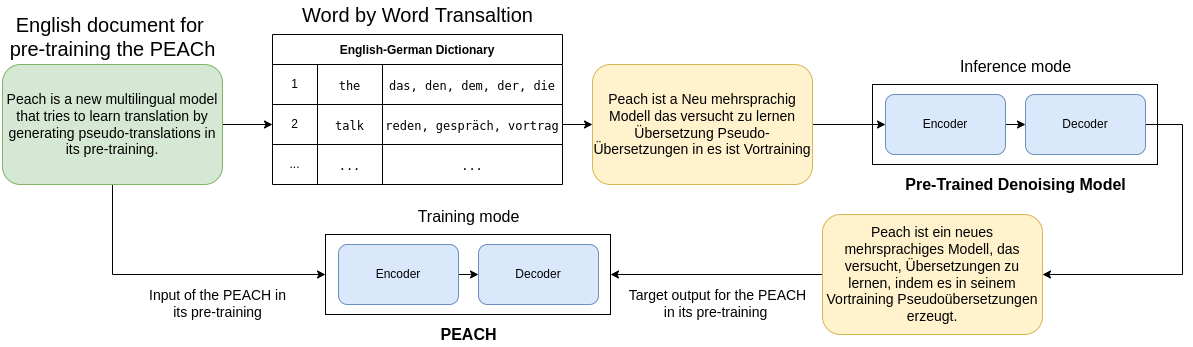}
    \caption{An overview of our pre-training pipeline for training a model based on SPDG. The method uses the output of the word-by-word translation of a pre-training document as the input of the trained denoising model based on Figure \ref{fig:denoising} to improve its quality.}
    \label{fig:word_by_word}
\end{figure*}

\section{PEACH}
PEACH is a new sequence-to-sequence  multilingual transformer model trained with SPDG, a semi-supervised pseudo-parallel document generation method. This section explains the pre-training objective and the model architecture.

\subsection{Semi-Supervised Pseudo-Parallel Document Generation (SPDG)}
\label{sec:SPDG}

Our proposed pre-training objective, SPDG, generates a pseudo-translation of the input document. For generating pseudo-translations, we use \citet{kim-etal-2018-improving}'s approach for unsupervised translation with some modifications. Our pipeline for pre-training a model based on SPDG is shown in Figure \ref{fig:word_by_word}. We pre-train a seq2seq denoising model for the target language using the pre-training corpus of that language. Next, for each pre-training document in the source language, we translate it to the target language word-by-word using dictionaries. Then, we give this word-by-word translated document to the pre-trained model with denoising objectives to improve its quality and restore missing words. 

Using this method, we can generate the pseudo-translation of each pre-training document from the source language to the target language. We use these pseudo-translations as gold translations for each pre-training document to pre-train a new language model for translation tasks. Since this pre-training objective is similar to translation, we hypothesize that the pre-trained model learns the translation task faster than the models trained using monolingual data. 

\paragraph{Word-by-Word Translation Using Dictionaries}

The first step to generate pseudo-parallel documents is to map sentences from one language to another using dictionaries. We used bilingual dictionaries provided by \citet{dictionaries} for our work. 
To map sentences word-by-word from one language to another, we first tokenize sentences using the NLTK\footnote{\url{https://www.nltk.org/}} library. Then, for each token, we find a translation for the token in the target language using a dictionary from the source to the target language. Some tokens, such as punctuations and numbers, do not need to be translated to the target language  because they are shared between them. Therefore, we just put them in the translated words set. Furthermore, we can not find any translation for named entities in dictionaries. To solve this issue, spaCy\footnote{\url{https://spacy.io/}} small (<lang>\_core\_news\_sm) models for named entity recognition for each language are used to extract named entities. We transliterate the named entities and put them in the translated words set. Tokens without translation in dictionaries that are not named entities, punctuations, or numbers are skipped. We hope denoising objectives could find an appropriate substitute for these tokens in the next step. The implementation details of word-by-word translation can be found in Appendix \ref{appendix:wbyw-imp}.

\paragraph{Improving Word-by-Word Translations with Denoising Objectives}

A critical problem with word-by-word translation is that the word order in the target language is not usually the same as the source. Furthermore, some words in the source language might not have any translation in the target language or vice versa. Additionally, since many words have multiple meanings, word-by-word translation might select the wrong translation for a word.

We define four denoising objectives to overcome the mentioned challenges, and train a denoising model for each language. Since Easy Data Augmentation (EDA) has shown great impact on the performance of various tasks in NLP \cite{wei2019eda,zhong2020random,abaskohi-etal-2022-utnlp}, our denoising objectives include word addition, word erasing, word substitution, and shuffling objectives of EDA. The pipeline is shown in Figure \ref{fig:denoising}. First, we shuffle the words in each sentence in a document while keeping the relative order of shuffled words in different sentences in the document. Next, we remove, add, and replace some of the words in each sentence to encourage the model to resolve the aforementioned issues in word-by-word translation. We use the corrupted document as the model's input and ask the model to generate the original one as its output.

The deshuffling objective aims to improve the ability of the model to reorder word-by-word translated documents. Removing and adding words help the model to correct some translations. Moreover, replacing is beneficial especially for correcting the word-by-word translation of ambiguous words.

Figure \ref{fig:word_by_word} depicts our pipeline for pre-training with SPDG on a single example. In the mentioned example, after word-by-word translation, some of the words in the pre-training document cannot be translated into German because they do not exist in the dictionary. Furthermore, the relative order of words in the word-by-word translated text is not grammatically correct, and some words can be substituted with more suitable ones. It can be seen that after applying the denoising model to the word-by-word translated text, the mentioned problems are resolved.


\subsection{Pre-Training with Multilingual SPDG}
\label{sec:MSPDG}

Most common multilingual models, such as mT5 \citep{mt5} and mBART \citep{mbart}, use MLM and MLM with Reordering as their pre-training objectives.
Despite their success, these objectives are not perfectly aligned with the goal of MT. Specifically, these objectives are designed to work on monolingual inputs; they denoise the input document in a specific language and produce the denoised version in the same language. 
Here, we design Algorithm \ref{algo:MSPDG}, in which the pre-training task's input is in one language, and its output is in another language. The algorithms' inputs are the corpora of all languages that the model should be trained on as well as their names. The algorithm generates the input-output pairs for pre-training the multilingual model.

\begin{algorithm} 
\small
\caption{Multilingual SPDG}
\label{algo:MSPDG}
\SetKwInOut{Input}{Input}
\SetKwInOut{Output}{Output}
\Input{Corpora, Langs}
\Output{MInputs,Moutputs}
\begin{algorithmic}
    \STATE $MInputs := \emptyset$
    \STATE $MOutputs := \emptyset$
    \FOR{$corpus$ \textbf{in} $Corpora$}
        \FOR{$doc$ \textbf{in} $corpus$}
            \FOR{$lang$ \textbf{in} $Langs - Lang(doc)$}
                \STATE $pst := SPDG(doc,Lang(doc),lang)$
                \STATE $MInputs := MInputs \cup \{doc\}$
                \STATE $MOutputs := MOutputs \cup \{pst\}$
            \ENDFOR
        \ENDFOR
    \ENDFOR
\end{algorithmic}
\end{algorithm}

In Algorithm \ref{algo:MSPDG}, given a pre-training document, we generate a pseudo-translation of it to each of the other languages. So, the model can observe translations in different languages for a single document. 
This helps the model in learning cross-lingual knowledge even about a language  not present in a specific training instance.
The mentioned claim is because the model learns about the language differences by translating the same input into multiple languages. 

It should be noted that based on the goal of pre-training a language model for translation, it is possible to change Algorithm \ref{algo:MSPDG}. For example, if the multilingual model is going to be used to just translate from or to English, there is no need to pre-train the model with the task of generating pseudo-translation from German to French. Since we are interested in evaluating our model on all pairs of the pre-training languages, we generate pseudo-translation for all pairs in Algorithm \ref{algo:MSPDG}. 

\paragraph{Architecture}

Our model, PEACH, and the other presented denoising models are all based on transformer \citep{transformer} encoder-decoder architecture with a 12 layer encoder and a 12 layer decoder with 768 hidden size, 3072 feed-forward filter size, and 12 self-attention heads.

\begin{table*}[!ht]
\small
\centering
\renewcommand{\arraystretch}{1.25}
\begin{tabular}{c|ccc}
\hline
\multirow{2}{*}{\textbf{Model}} & \multicolumn{2}{c}{\textbf{WMT14}} & \textbf{
WMT19}\\
& \textbf{FR$\xleftrightarrow{}$EN} & \textbf{DE$\xleftrightarrow{}$EN} & \textbf{DE$\xleftrightarrow{}$FR}\\
\hline
\textbf{MLM} & $21.38 \xleftrightarrow{} 21.64$ & $17.88 \xleftrightarrow{} 19.54$ & $16.59 \xleftrightarrow{} 16.54$ \\
\textbf{MLM with Reordering} & $29.02 \xleftrightarrow{} 28.71$ & $22.80 \xleftrightarrow{} 25.53$ & $21.39 \xleftrightarrow{} 22.45$ \\
\textbf{Transformer} & $9.15 \xleftrightarrow{} 9.17$ & $10.02 \xleftrightarrow{} 9.79$ & $9.16 \xleftrightarrow{} 10.31$ \\
\hline
\textbf{PEACH} & $\textbf{31.25} \xleftrightarrow{} \textbf{29.98}$ & $\textbf{23.61} \xleftrightarrow{} \textbf{26.97}$ & $\textbf{23.13} \xleftrightarrow{} \textbf{25.25}$ \\
\hline
\end{tabular}
\caption{The supervised translation results evaluated with BLEU score. }
\label{table:supervised-results-tranlation}
\end{table*}

\section{Experiments}

In this section, we compare the results of PEACH, trained with SPDG, with other common objectives utilized for pre-training multilingual models. To investigate the effectiveness of SPDG in comparison with common objectives, we pre-trained two other models based on mT5's MLM objective \citep{mt5}  and mBART's MLM with Reordering objective \citep{mbart} in the same setup.

The codes for pre-training and fine-tuning of all models are publicly available on GitHub\footnote{\url{https://github.com/AmirAbaskohi/PEACH}}.

\subsection{Pre-Training Data and Configuration}

We pre-train PEACH on English, French, and German with the CC100 corpora \cite{cc100-1,cc100-2}. Due to the lack of computing power, we cannot use more than around $550M$ words of text from each language. So, we train our model on around $1.6B$ total words. Our pre-training batch size is 96, with a maximum of 512 input and output tokens, and we train it for 500K steps on Google Colab TPUs (v2-8). The AdaFactor \citep{adafactor} optimizer with a decay rate of 0.8 and a dropout rate of 0.1 is used in pre-training and fine-tuning. Furthermore, we use the SentencePiece BPE algorithm \citep{bpe,Kudo2018SentencePieceAS} to generate a vocabulary of 32K words for denoising models and 96k for multilingual models. We pre-train PEACH with Multilingual SPDG for 75\% of its pre-training steps and mT5's MLM \citep{mt5} approach for the other 25\% pre-training steps. The latter pre-training objective is used because it increases the scope of the fine-tuning tasks that our model can do well. Indeed, multilingual SPDG teaches the model to transform a text from one language to another, but it does not help the model in tasks where their inputs and outputs are in the same language. Therefore, pre-training the model with MLM for a few steps is helpful.

We train the denoising models with the same setup as PEACH. An important factor in training denoising models is the rate of corruption for training documents. We shuffle all words in sentences while removing, adding, and replacing a small proportion of them. We use the word-by-word translation script outputs to decide on these rates. First, we calculate the rate of missing words in word-by-word translation using dictionaries for all languages to a specific language on around 1GB of text of each language. Then, we use a normal distribution with mean and standard deviation of the same as the calculated numbers to define the rate of words that should be removed from a sentence. The values of corruption rates for each language are reported in Table \ref{appendix:table:denosing-rate} in Appendix \ref{appendix:denoising-details}, in which we explain the method to find the best values for rates. 

Due to the lack of computing power, we cannot train a large-scale PEACH and compare it with pre-trained models like mT5 or mBART. Instead, we train two models based on mT5 \citep{mt5} objective, which we call MLM, and mBART \citep{mbart} objective, which we call MLM with Reordering, with the same setup as PEACH. Also, we fine-tune a Transformer model with randomly initialized weights on downstream tasks. 

\subsection{Results} 
This section evaluates PEACH in various translation scenarios, including supervised, zero- and few-shot. We also evaluate PEACH's ability for cross-lingual knowledge transfer in translation and natural language inference tasks.

\begin{table*}[!ht]

\centering
\renewcommand{\arraystretch}{1.25}
\begin{tabular}{c|ccc}
\hline
\multirow{2}{*}{\textbf{Model}} & \multicolumn{2}{c}{\textbf{WMT14}} & \textbf{
WMT19}\\
& \textbf{FR$\xleftrightarrow{}$EN} & \textbf{DE$\xleftrightarrow{}$EN} & \textbf{DE$\xleftrightarrow{}$FR}\\
\hline
\textbf{SPDG\textsubscript{$EN\xleftrightarrow{}FR$} (200k steps)} & $25.98 \xleftrightarrow{} 25.42$ & $-$ & $-$ \\
\textbf{SPDG\textsubscript{$EN\xleftrightarrow{}DE$} (200k steps)} & $-$ & $17.75 \xleftrightarrow{} 22.97$ & $-$ \\
\textbf{SPDG\textsubscript{$FR\xleftrightarrow{}DE$} (200k steps)} & $-$ & $-$ & $16.24 \xleftrightarrow{} 18.77$ \\
\hline
\textbf{SPDG\textsubscript{$EN\xleftrightarrow{}FR\xleftrightarrow{}DE$} (100k steps)} & ${27.40} \xleftrightarrow{} {26.60}$ & ${21.21} \xleftrightarrow{} {23.89}$ & ${20.49} \xleftrightarrow{} {22.32}$ \\
\textbf{SPDG\textsubscript{$EN\xleftrightarrow{}FR\xleftrightarrow{}DE$} (200k steps)} & $\textbf{29.04} \xleftrightarrow{} \textbf{28.08}$ & $\textbf{22.33} \xleftrightarrow{} \textbf{25.29}$ & $\textbf{21.67} \xleftrightarrow{} \textbf{23.29}$ \\
\hline
\end{tabular}
\caption{Results of different models trained with SPDG on either two or three indicated languages. The number of pre-training steps is shown in parenthesis.}
\label{table:peach-vs-pairs}
\end{table*}

\begin{figure}[!ht]
    \centering
    \includegraphics[scale=0.36]{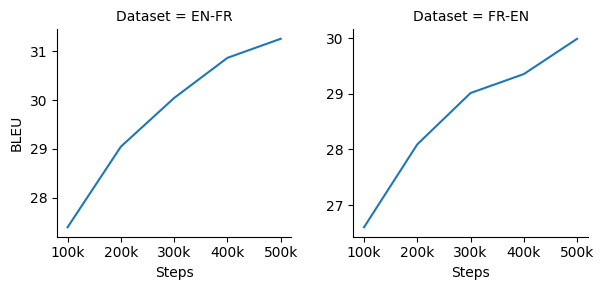}
    \caption{PEACH's performance in pre-training steps on WMT14's EN-FR section. Results for EN-DE and DE-FR are reported in Table \ref{appendix:table:pretraining-performance} in  Appendix \ref{appendix:figure-details}.}
    \label{fig:tranlation-supervised-steps-results}
\end{figure}

\paragraph{Supervised Translation}

In order to evaluate PEACH on translation tasks, we fine-tune it on the EN-DE and EN-FR parts of the WMT14 dataset \citep{wmt14}. Additionally, we fine-tune our model on the FR-DE part of the WMT19 dataset \citep{wmt19} in the same setup. Since the test set of WMT19 DE-FR datasets is not available publicly to the best of our knowledge, we evaluated the models on its validation set. The model is fine-tuned for 50K steps with a batch size of 96, a learning rate of $5\times10^{-5}$, and the same optimizer as pre-training. We use 10K warmup steps for fine-tuning. More information about the experiments' setup is reported in Appendix \ref{appendix:fine-tuning-details}. It should be noted that while translation downstream datasets usually have millions of samples, we at most use $50000\times96$ samples of them due to the lack of computing power. To support the selected number of samples for the downstream task, we report pre-training and fine-tuning time on the whole datasets for an epoch in Appendix \ref{appendix:fine-tuning-details}. This sample count is less than 15\% of samples for the WMT14 English-French dataset. Additionally, since the primary purpose of this paper is to introduce a new method for pre-training multilingual models and the comparisons happen in the same setup for all objectives, the results are fair and valid.

The results of our model and other trained models on translation tasks are reported in Table \ref{table:supervised-results-tranlation}. 
Additionally, the results of our model on EN-FR downstream dataset in some pre-training steps are shown in Figure \ref{fig:tranlation-supervised-steps-results}. Also, the results for other downstream datasets are reported in Table \ref{appendix:table:pretraining-performance} in Appendix \ref{appendix:figure-details}. The presented results show that PEACH outperforms other models, not only with 500K steps of pre-training but also even with its 200K steps pre-training checkpoint. 
Furthermore, the MLM method used in mT5 achieves worse results than MLM with Reordering objective that mBART used. We believe this is because the MLM objective of mT5 just asks the model to generate the masked spans in the output, while mBART's objective asks the model to reorder and predict the masked spans of the input document simultaneously. Indeed, the objective of mBART asks the model to generate complete sentences in its output, and that is why it can generate better translations. On the other hand, mT5 just predicts spans of text, which are not complete sentences in many cases. 

We believe that the better results of our model stem from its pre-training objective which is similar to translation tasks. Indeed, we pre-trained our model on a massive amount of pre-training data with a task similar to translation, which increases the model's ability in translation when it is fine-tuned with a smaller amount of translation samples.

To investigate the effect of pre-training on more than two languages on the performance of our model on translation tasks, we pre-train a model based on SPDG for 200K steps  for each pair of languages, and fine-tune them for 50k steps, with the same setup as PEACH. The results are reported in Table \ref{table:peach-vs-pairs}. We show that our multilingual model with three languages outperforms other models not only with full pre-training for 200K steps but also with 100K steps of pre-training. We believe this is because we perform the SPDG objective between each pair of languages in its pre-training. Indeed, this approach for pre-training multilingual models helps the model simultaneously gain knowledge about other languages than the pair of languages in each pre-training example because it observes the same input with different outputs for each language. These results support our claim in section \ref{sec:MSPDG}.

\paragraph{Zero- and Few-Shot Translation}

\begin{figure}[!ht]
    \centering
    \includegraphics[scale=0.31]{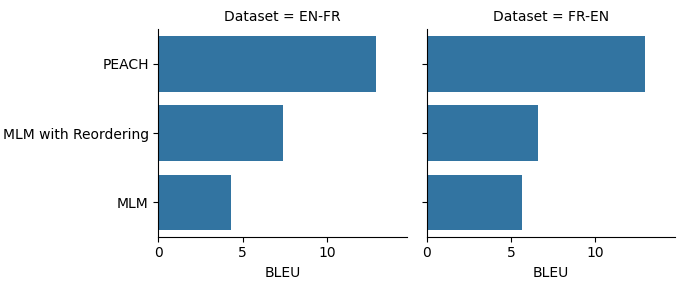}
    \caption{Comparing the pre-trained models in zero-shot setting on WMT14 EN-FR section. Results for EN-DE and DE-FR are reported in Table \ref{appendix:table:zero-results} in  Appendix \ref{appendix:figure-details}.}
    \label{fig:zero-experiment}
\end{figure}

We evaluate the pre-trained models in a zero-shot setting to investigate our model's ability in low-resource scenarios. Each pre-trained model is evaluated on the test set of WMT14 EN-FR dataset without fine-tuning. The results of this experiment are reported in Figure \ref{fig:zero-experiment}. The results for EN-DE and DE-FR section of WMT14 and WMT19 are reported in Table \ref{appendix:table:zero-results} in Appendix \ref{appendix:figure-details}. The results in Figure \ref{fig:zero-experiment} and Table \ref{appendix:table:zero-results} show that our model, PEACH, outperforms other models in zero-shot translation.
We believe this stems from the similarity of its pre-training objective with actual translation tasks.

\begin{figure}[!ht]
    \centering
    \includegraphics[scale=0.57]{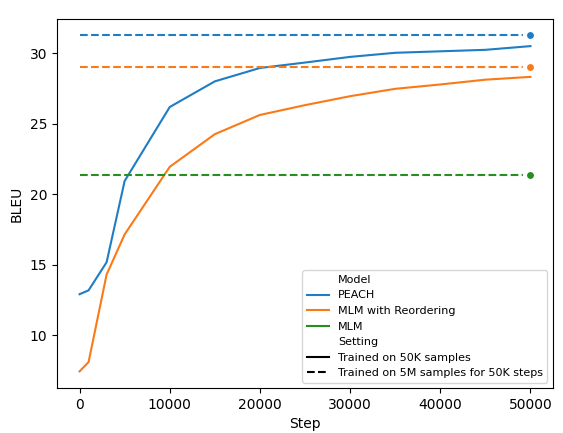}
    \caption{Results of fine-tuning PEACH with 50K samples of WMT14 EN-FR dataset for 0 to 50k steps, and its comparison with MLM and MLM with Reordering objectives on $50000\times96$ data points. PEACH outperforms the fully-trained MLM models after only 25K fine-tuning steps.}
    \label{fig:zero-few-experiment-steps}
\end{figure}

For few-shot experiments, we fine-tuned PEACH on 50K samples from the English-French section of the WMT14 dataset at a maximum of 50K steps. The results are shown in Figure \ref{fig:zero-few-experiment-steps}. Accordingly, PEACH outperforms MLM with Reordering model trained in the same setup. Additionally, PEACH surpasses MLM and MLM with Reordeing models' checkpoints in 50K fine-tuning steps on around 5M samples, after only 10K and 25K steps of fine-tuning on 50K samples. We conclude that PEACH performs well in low-resource scenarios because it is trained on a massive amount of psuedo-translation data.

\begin{table*}[!ht]
\scriptsize
\centering
\renewcommand{\arraystretch}{1.25}
\begin{tabular}{c|ccc|ccc}
\hline
\multirow{3}{*}{\textbf{Fine-Tuned / Evaluated}} & 
\multicolumn{3}{c}{\textbf{PEACH}} & \multicolumn{3}{c}{\textbf{
MLM with Reordering}}\\
& \multicolumn{2}{c}{\textbf{WMT14}} & \textbf{
WMT19} & \multicolumn{2}{c}{\textbf{WMT14}} & \textbf{
WMT19}\\
& \textbf{FR$\xleftrightarrow{}$EN} & \textbf{DE$\xleftrightarrow{}$EN} & \textbf{DE$\xleftrightarrow{}$FR} & \textbf{FR$\xleftrightarrow{}$EN} & \textbf{DE$\xleftrightarrow{}$EN} & \textbf{DE$\xleftrightarrow{}$FR}\\
\hline
\textbf{$EN \xrightarrow{} FR$} & $-  \xleftrightarrow{} 11.38$ & $12.35 \xleftrightarrow{} 16.57$ & $12.38 \xleftrightarrow{} 21.91$ & $-  \xleftrightarrow{} 11.25$ & $11.52 \xleftrightarrow{} 12.51$ & $11.65 \xleftrightarrow{} 11.70$ \\
\textbf{$FR \xrightarrow{} EN$} & $11.30 \xleftrightarrow{} -$ & $14.62 \xleftrightarrow{} 21.35$ & $15.05 \xleftrightarrow{} 17.28$ & $11.27 \xleftrightarrow{} -$ & $12.88 \xleftrightarrow{} 12.99$ & $12.68 \xleftrightarrow{} 11.28$ \\
\textbf{$EN \xrightarrow{} DE$} & $20.63 \xleftrightarrow{} 11.99$ & $- \xleftrightarrow{} 12.70$ & $19.88 \xleftrightarrow{} 13.84$ & $10.80 \xleftrightarrow{} 11.29$ & $- \xleftrightarrow{} 12.64$ & $12.89 \xleftrightarrow{} 11.07$ \\
\textbf{$DE \xrightarrow{} EN$} & $18.97 \xleftrightarrow{} 24.54$ & $13.39 \xleftrightarrow{} -$ & $14.99 \xleftrightarrow{} 18.85$ & $10.99 \xleftrightarrow{} 13.85$ & $12.71 \xleftrightarrow{} -$ & $11.23 \xleftrightarrow{} 11.09$ \\
\textbf{$FR \xrightarrow{} DE$} & $23.64 \xleftrightarrow{} 24.69$ & $18.59 \xleftrightarrow{} 22.69$ & $- \xleftrightarrow{} 23.35$ & $12.07 \xleftrightarrow{} 11.43$ & $12.81 \xleftrightarrow{} 11.54$ & $- \xleftrightarrow{} 20.65$ \\
\textbf{$DE \xrightarrow{} FR$} & $24.88 \xleftrightarrow{} 24.94$ & $20.12 \xleftrightarrow{} 20.74$ & $23.03 \xleftrightarrow{} -$  & $14.72 \xleftrightarrow{} 11.57$ & $12.86 \xleftrightarrow{} 11.92$ & $21.56 \xleftrightarrow{} -$ \\
\hline
\end{tabular}
\caption{The results of experiments on cross-lingual knowledge transfer for translation. We fine-tune the model on one language and evaluate it on other languages. The results are reported using BLEU score.}
\label{table:cross-lingual-results-tranlation}
\end{table*}

\begin{table}[!ht]
\small
\centering
\renewcommand{\arraystretch}{1.2}
\begin{tabular}{c|ccc}
\hline
\multirow{2}{*}{\textbf{Model}} & \multicolumn{3}{c}{\textbf{XNLI}} \\
& \textbf{EN} & \textbf{FR} & \textbf{DE}\\
\hline
\textbf{MLM} & .676 & .480 & .463\\
\textbf{MLM with Reordering} &  .710 & .603 & .527\\
\hline 
\textbf{PEACH} & \textbf{.745} & \textbf{.637} & \textbf{.636}\\
\hline
\end{tabular}
\caption{The accuracy results on the XNLI benchmark.}
\label{table:xnli}
\end{table}




\paragraph{Cross-Lingual Transfer for Translation}

Here we evaluate each fine-tuned model on a language pair on how it performs for other pairs and directions. We use the fine-tuned models in Table \ref{table:supervised-results-tranlation} for these experiments.

The experimental results in Table \ref{table:cross-lingual-results-tranlation} demonstrate that PEACH can transfer the knowledge learned from one language pair to another better than MLM with Reordering model. We believe this stems from our pre-training method in which we ask the model to generate pseudo-translations between each pair of languages.
Furthermore, the results confirm \citet{mbart}'s experiments and show that whenever a model fine-tuned on a dataset from A to B is evaluated on A to C or C to B or B to A, the results on the evaluation dataset increase more than other combinations. Additionally, because the inputs of PEACH's encoder are human-generated texts while the decoder's expected outputs are the outputs of the denoising models, fine-tuning from A to B increases the performance of C to B more than A to C. Indeed, fine-tuning from A to B helps the decoder of our model learn to generate better outputs by observing human-generated texts in its decoder. This is because our model did not encounter human-generated texts as gold labels in its output during pre-training. On the other hand, observing more human-generated inputs is not as helpful as human-generated outputs since the inputs of the model's encoder were human-generated text in its pre-training. 

In support of the previous point, the results in Table \ref{table:cross-lingual-results-tranlation} show that PEACH fine-tuned on the DE-EN dataset achieves better results than MLM fine-tuned on the FR-EN dataset, when evaluated on the FR-EN dataset. Additionally, PEACH fine-tuned on the EN-FR dataset achieves a comparable result with MLM with Reordering fine-tuned on the DE-FR dataset, when evaluated on the DE-FR dataset (0.54 difference in BLEU). We believe this experiment shows PEACH's ability to transfer the knowledge learned from a language to another effectively.

\paragraph{Cross-Lingual Transfer for natural language inference}
We focus on translation in this paper. However, we expect that PEACH's ability to transfer knowledge between languages is suitable for other cross-lingual scenarios as well.
To test this hypothesis, we evaluate PEACH on the XNLI benchmark \citep{xnli}. 
We fine-tune our model for 50K steps with a batch size of 256, a learning rate of $10^{-3}$, and a maximum output length of 16 on the MultiNLI English dataset \citep{multinli} and apply it to the XNLI benchmark. The results of this experiment are reported in Table \ref{table:xnli}. 

According to Table \ref{table:xnli}, PEACH outperforms other models in transferring knowledge from English to German and French. Considering our pre-training objective, in which we ask the model to generate pseudo-translations for each pair of pre-training languages, we believe this objective helps PEACH to transfer the knowledge about the English dataset to other languages better than other pre-trained models.

\section{Conclusion}

We introduced SPDG, a semi-supervised method for pre-training multilingual seq2seq models, to address the lack of parallel data between different languages. In this new method, we use bilingual dictionaries and denoising models trained with reordering, adding, substituting, and removing words to generate a pseudo-translation for each pre-training document. We use this generated data to train our multilingual model, PEACH, for English, French, and German languages. Our results show that PEACH outperforms the common pre-training objectives for training multilingual models. Furthermore, PEACH shows a remarkable ability in zero- and few-shot translation and knowledge transfer between languages.

\section*{Limitations}

The main limitations of our work can be classified into two types: 1) SPDG's limitations and 2) Computational limitations.

\paragraph{SPDG's Limitations} Although our method can address the issue of limited parallel data between different languages, it does not solve the problem completely. First, our method uses bilingual dictionaries to translate each pre-training document from one language to another, which is not always available for low-resource languages. Furthermore, the available dictionaries for low-resource languages do not have a high quality and are not comparable with high-resource languages. Additionally, we use Named Entity Recognition (NER) models to transfer named entities of each pre-training document into its pseudo-translation, which is unavailable for some low-resource languages. Therefore, using unsupervised methods for NER can be a solution for the mentioned problem, which is not investigated in this work. 

\paragraph{Computational limitations}
We did not have access to clusters of GPU or TPU to train our models on a large scale and compare them with the results reported in other papers about multilingual models. However, we tried to provide a realistic setting for our experiments. Further investigation into training models on a larger scale, same as standard multilingual models, can improve this work. 


\bibliography{anthology,custom}
\bibliographystyle{acl_natbib}

\appendix

\section{Pre-Training and Downstream Datasets' Information}

We use the CC100 corpus \citep{cc100-1, cc100-2} for pre-training all models in this work. More specifically, we used the English (EN), French (FR), and German (DE) parts of the mentioned corpus. Due to the lack of computing power and the massive amount of paragraphs in this corpus, we use around 3GB of the text of each language, approximately 550M words from each language and a total of 1.6B words, to pre-train our models. For more reproducibility, we select $500000 \times 96$, the total pre-training steps multiplied by the used batch size, first paragraphs of each mentioned language in the corpus as pre-training samples.

In order to evaluate the models for translation tasks, we use English to French and English to German sections of the WMT14 \citep{wmt14} and the French to German part of the WMT19 dataset \citep{wmt19}. The total number of samples in each set of each pre-training dataset is reported in Table \ref{appendix:table:downstream-info}. We do not use all the samples in all datasets due to the lack of computing power. We use at most $50000 \times 96$, the total fine-tuning steps multiplied by the batch size, unique samples of each dataset. We use all the samples for datasets with fewer samples than the mentioned number. To the best of our knowledge, the test set of the WMT19 FR-DE dataset is not publicly available. Therefore, we report the results on the validation set instead of the test set.

For the experiment on transferring knowledge from one language to another, we fine-tune PEACH on the MultiNLI dataset \citep{multinli}, consisting of natural language inference samples for the English language. Then, we evaluate the model on English, French, and German samples in test set in the XNLI dataset \citep{xnli}, consisting of natural language inference samples for the mentioned languages. The number of samples for each dataset is reported in Table \ref{appendix:table:nli-samples}. Both mentioned datasets use three labels; neutral, entailment, and contradiction.

\begin{table}[!ht]
\small
\centering
\renewcommand{\arraystretch}{1.2}
\begin{tabular}{c|cccc}
\hline
\textbf{Dataset} & \textbf{Language} & \textbf{Train} & \textbf{Dev} & \textbf{Test} \\
\hline
WMT14 & EN$\xleftrightarrow{}$FR & 40836715 & 3000 & 3003 \\ 
WMT14 & EN$\xleftrightarrow{}$DE & 4508785 & 3000 & 3003 \\ 
WMT19 & EN$\xleftrightarrow{}$FR & 9824476 & 1512 & - \\ 
\end{tabular}
\caption{Number of Samples in supervised stranslation datasets.}
\label{appendix:table:downstream-info}
\end{table}

\begin{table}[!ht]
\small
\centering
\renewcommand{\arraystretch}{1.2}
\begin{tabular}{c|ccc}
\hline
\textbf{Dataset} & \textbf{Language} & \textbf{Train} & \textbf{Test} \\
\hline
MultiNLI & EN & 392702 & - \\ 
XNLI & EN & - & 5010 \\ 
XNLI & DE & - & 5010 \\ 
XNLI & FR & - & 5010 \\ 
\end{tabular}
\caption{Number of Samples in natural language inference datasets.}
\label{appendix:table:nli-samples}
\end{table}

\section{Word-by-Word Translation Implementation Details}
\label{appendix:wbyw-imp}
The word-by-word translation was performed in batches of 1K documents. The batch size does not affect the algorithm's performance and should be chosen based on the available resources. 

After lower casing the documents in a batch, named entities are extracted using the spaCy toolkit. The identified entities should be divided by white space characters since the named entities sometimes consist of multiple words. Since the spaCy toolkit for named entity recognition sometimes chooses definite articles as a part of named entities, we filter out definite articles such as "the," "le," "la," "les," "der," "die," and "das" and translate them using dictionaries in following steps.

In order to perform word-by-word translation, we first tokenize the document. We search for the translation of each token from the source language to the destination language using the appropriate dictionary. If we found more than one possible translation for a token, we uniformly select one of them. Suppose we can not find any translations for a token in the source to the destination language dictionary. In that case, we use source to English and English to destination dictionary to find a translation for the mentioned token. First, we search for a translation from the source language to English using the source to English dictionary. Next, we search for a translation from English to the destination language in the English to the destination dictionary. This technique is just helpful when there is a translation from the source token to English. If we can not find any translations for a token, we mark it as unknown to decide about it later.

For the terms that were marked as unknown, if the token contains numbers or punctuations, we transfer it without any change to the output as a translated word. Otherwise, we check if the word is in the extracted named entities. In this case, we transliterate the word into the destination language using polyglot library \footnote{\url{https://polyglot.readthedocs.io/en/latest/Transliteration.html}} and put it in the output as a translated word. For complex words such as "high-end," we break the word into its alphabetical components and search them in the dictionary. If we could find a translation for all components, we would translate each component and concatenate them using the proper separator. In the case that none of the aforementioned scenarios happens, we omit the word and hope the denoising pre-trained model can find a proper translation for it.

\section{Denoising Models Pre-Training and Corruption Rate Details}
\label{appendix:denoising-details}

\begin{table}[!ht]
\small
\centering
\renewcommand{\arraystretch}{1.2}
\begin{tabular}{c|ccc}
\hline
\textbf{Language} & \textbf{Removing} & \textbf{Addition} & \textbf{Substitution} \\
\hline
EN & .066/.061 & .01-.03 & .05-.07 \\ 
FR & .152/.087 & .01-.03 & .05-.07 \\ 
DE & .137/.085 & .01-.03 & .05-.07 \\ 
\end{tabular}
\caption{Rates used for pre-training objectives of Denoising models. For removing, we report mean/std.}
\label{appendix:table:denosing-rate}
\end{table}

The procedure for generating pre-training data for training the denoising model is shown in Figure \ref{fig:denoising}. This procedure consists of sentence shuffling, word removing, addition, and substitution.

The first step for generating pre-training data is loading a batch of pre-training documents into the memory as the current batch. We used a batch size of 1K for generating pre-training data for training denoising models. The batch size plays an essential role in this procedure because our algorithm selects candidates for replacing some words in a sample from the words available in other sample in the current batch. We did not investigate the effect of batch size due to the lack of computing power.

After tokenizing the separated sentences using the NLTK toolkit, we shuffle the words in each sentence but keep the relative order of sentences. It helps the denoising model learn the relative order of sentences, which is crucial since the word-by-word translation algorithm might face documents with multiple sentences. Therefore, this will teach the denoising model how to figure out the boundaries of different sentences.

Next, for each sentence, we select $m \times c$ words to be replaced, in which $m$ is the length of the sentence and $c$ is a random number from a uniform distribution between the reported rates in Table \ref{appendix:table:denosing-rate}. The algorithm selects $m \times c$ unique words from other samples in the current batch uniformly to be substituted with the selected words of the current document. The word addition objective works the same way as the substitution, but the algorithm does not replace any words. The word removing objective works the same, but it uses a normal distribution for generating the random number, and it just omits some words from each sentence without replacing them with other words.

The word substitution and addition rates in Table \ref{appendix:table:denosing-rate} were selected based on observation of the outputs of the word-by-word translation algorithm. On the other hand, we computed the mean and standard deviation for the proportion of words that the word-by-word translation algorithm could not find any translation for them on the pre-training corpus. The main purpose of the word removing objective is to find a translation for the words that the word-by-word translation algorithm could not find any translation for them by considering the context of the sentence. Therefore, computing this number on the pre-training corpus that the final multilingual model will be trained on will improve the denoising model's ability to denoise the word-by-word translation algorithm's outputs. This decreases the number of words that the word-by-word translation algorithm or the denoising model could not find a translation for them.

\section{Experiment Details and Setup}
\label{appendix:fine-tuning-details}

It takes six days to pre-train PEACH on $500000\times96$ pre-training documents for 500K steps and a batch size of 96. However, the downstream dataset for English-French translation consists of almost 40M samples, which is only 8M less than our pre-training documents and takes five days to fine-tune for just one epoch. Therefore, choosing $50000\times96$ samples for fine-tuning is plausible due to the number of total pre-training documents and steps of the model's pre-training. The setup for training denoising models is reported in Table \ref{appendix:table:pretrain-setting-denoising}. The experiment setups for pre-training multilingual models on English, French, and German are reported in Table \ref{appendix:table:pretrain-setting-multi}. The setups for pre-training bilingual models used in different experiments are reported in Table \ref{appendix:table:pretrain-setting-bi}. Table \ref{appendix:table:fine-tuning-details} reports the details of experiments on fine-tuning models on supervised translation tasks. The experiment setup for the few-shot scenario is reported in Table \ref{appendix:table:fewshot}. The experiment setup for the fine-tuning on the XNLI \citep{xnli} task is reported in Table \ref{appendix:table:nli}.

\begin{table*}
\centering
\small
\renewcommand{\arraystretch}{1.3}
\begin{tabular}{c|ccccc}
\hline
\textbf{Language} & \textbf{Learning rate} & \textbf{Steps} & \textbf{Batch Size} & \textbf{Max Input Length} & \textbf{Max Output Length} \\
\hline
English & .01 & 500K & 96 & 512 & 512 \\
German & .01 & 500K & 96 & 512 & 512 \\
French & .01 & 500K & 96 & 512 & 512 \\
\hline
\end{tabular}
\caption{\label{appendix:table:pretrain-setting-denoising} Pre-training settings for denoising models.}
\end{table*}

\begin{table*}
\centering
\scriptsize
\renewcommand{\arraystretch}{1.3}
\begin{tabular}{c|cccccc}
\hline
\textbf{Objective} & \textbf{Learning rate} & \textbf{Steps} & \textbf{Batch Size} & \textbf{Language} & \textbf{Max Input Length} & \textbf{Max Output Length} \\
\hline
Multilingual SPDG & .01 & 500K & 96 & EN-FR-DE & 512 & 512 \\
MLM with Reordering & .01 & 500K & 96 & EN-FR-DE & 512 & 512 \\
MLM & .01 & 500K & 96 & EN-FR-DE & 512 & 512 \\
\hline
\end{tabular}
\caption{\label{appendix:table:pretrain-setting-multi} Pre-training settings for multilingual models trained on English, German, and French.}
\end{table*}

\begin{table*}
\centering
\scriptsize
\renewcommand{\arraystretch}{1.3}
\begin{tabular}{c|cccccc}
\hline
\textbf{Objective} & \textbf{Learning rate} & \textbf{Steps} & \textbf{Batch Size} & \textbf{Language} & \textbf{Max Input Length} & \textbf{Max Output Length} \\
\hline
SPDG & .01 & 200K & 96 & EN-FR & 512 & 512 \\
SPDG & .01 & 200K & 96 & EN-DE & 512 & 512 \\
SPDG & .01 & 200K & 96 & DE-FR & 512 & 512 \\
\hline
\end{tabular}
\caption{\label{appendix:table:pretrain-setting-bi} Pre-training settings for bilingual models.}
\end{table*}

\begin{table*}
\centering
\scriptsize
\renewcommand{\arraystretch}{1.3}
\begin{tabular}{c|ccccccc}
\hline
\textbf{Dataset} & \textbf{Learning rate} & \textbf{Steps} & \textbf{Batch Size} & \textbf{Beam Size} & \textbf{Beam alpha} & \textbf{Max Input} & \textbf{Max Output} \\
\hline
WMT14 \textsubscript{EN-FR} & $5\times 10^{-5}$ & 50K & 96 & 1 & .6 & 512 & 512 \\
WMT14 \textsubscript{EN-DE} & $5\times 10^{-5}$ & 50K & 96 & 1 & .6 & 512 & 512 \\
WMT19 \textsubscript{DE-FR} & $5\times 10^{-5}$ & 50K & 96 & 1 & .6 & 512 & 512 \\
\hline
\end{tabular}
\caption{\label{appendix:table:fine-tuning-details} Fine-tuning settings for models used in supervised translation experiments.}
\end{table*}

\begin{table*}
\centering
\scriptsize
\renewcommand{\arraystretch}{1.3}
\begin{tabular}{c|cccccccc}
\hline
\textbf{Dataset} & \textbf{Learning rate} & \textbf{Sample count} & \textbf{Steps} & \textbf{Batch Size} & \textbf{Beam Size} & \textbf{Beam alpha} & \textbf{Max Input} & \textbf{Max Output} \\
\hline
WMT14 \textsubscript{EN-FR} & $5\times 10^{-5}$ & 50K & 1K-50K & 96 & 1 & .6 & 512 & 512 \\
\hline
\end{tabular}
\caption{\label{appendix:table:fewshot} Fine-tuning settings for the few-shot supervised translation experiment.}
\end{table*}

\begin{table*}
\centering
\scriptsize
\renewcommand{\arraystretch}{1.3}
\begin{tabular}{c|cccccccc}
\hline
\textbf{Dataset} & \textbf{Learning rate} & \textbf{Steps} & \textbf{Batch Size} & \textbf{Beam Size} & \textbf{Beam alpha} & \textbf{Max Input} & \textbf{Max Output} & \textbf{Language} \\
\hline
XNLI & $1\times 10^{-3}$ & 50K & 256 & 1 & .6 & 512 & 16 & EN-FR-DE \\
\hline
\end{tabular}
\caption{\label{appendix:table:nli} Fine-tuning settings for knowledge transfer experiment on natural language inference.}
\end{table*}

\section{Figures' Details and Information}
\label{appendix:figure-details}

The reported numbers in Figures \ref{fig:tranlation-supervised-steps-results}, \ref{fig:zero-experiment}, and \ref{fig:zero-few-experiment-steps}  are reported in Tables \ref{appendix:table:pretraining-performance}, \ref{appendix:table:zero-results}, and \ref{appendix:table:few-results} for better readability.

\begin{table*}
\centering
\scriptsize
\renewcommand{\arraystretch}{1.3}
\begin{tabular}{c|ccccc}
\hline
\textbf{Dataset // Pre-Training steps} & \textbf{100K steps} & \textbf{200K steps} & \textbf{300K steps} & \textbf{400K steps} & \textbf{500K steps} \\
\hline

WMT14 \textsubscript{FR $\xleftrightarrow{}$ EN} & $27.40\xleftrightarrow{}26.60$ & $29.04\xleftrightarrow{}28.08$ & $30.04\xleftrightarrow{}29.01$ & $30.86\xleftrightarrow{}29.35$ & $31.25\xleftrightarrow{}29.98$ \\
WMT14 \textsubscript{DE $\xleftrightarrow{}$ EN} & $21.21\xleftrightarrow{}23.89$ & $22.33\xleftrightarrow{}25.29$ & $22.87\xleftrightarrow{}26.07$ & $23.25\xleftrightarrow{}26.52$ & $23.61\xleftrightarrow{}26.97$ \\
WMT19 \textsubscript{DE $\xleftrightarrow{}$ FR} & $20.49\xleftrightarrow{}22.32$ & $21.67\xleftrightarrow{}23.29$ & $22.12\xleftrightarrow{}24.07$ & $22.65\xleftrightarrow{}24.70$ & $23.13\xleftrightarrow{}25.25$ \\
\hline
\end{tabular}
\caption{\label{appendix:table:pretraining-performance} PEACH's performance in different pre-training steps on downstream tasks evaluated with BLEU score. The fine-tuning setup is reported in Table \ref{appendix:table:fine-tuning-details}. These numbers are reported in Figure \ref{fig:tranlation-supervised-steps-results}.}
\end{table*}

\begin{table*}
\small
\centering
\renewcommand{\arraystretch}{1.25}
\begin{tabular}{c|ccc}
\hline
\multirow{2}{*}{\textbf{Model}} & \multicolumn{2}{c}{\textbf{WMT14}} & \textbf{
WMT19}\\
& {FR$\xleftrightarrow{}$EN} & {DE$\xleftrightarrow{}$EN} & {DE$\xleftrightarrow{}$FR}\\
\hline
{MLM} & $4.33 \xleftrightarrow{} 5.64$ & $6.40 \xleftrightarrow{} 5.69$ & $6.39 \xleftrightarrow{} 4.56$ \\
{MLM with Reordering} & $7.42 \xleftrightarrow{} 6.63$ & $7.73 \xleftrightarrow{} 7.96$ & $7.17 \xleftrightarrow{} 7.71$ \\
{PEACH} & $12.89 \xleftrightarrow{} 12.98$ & $11.75 \xleftrightarrow{} 14.05$ & $11.83 \xleftrightarrow{} 13.11$ \\
\hline
\end{tabular}
\caption{The zero-shot translation results of the models evaluated with BLEU score. These numbers are reported in Figure \ref{fig:zero-experiment}}
\label{appendix:table:zero-results}
\end{table*}

\begin{table*}
\centering
\renewcommand{\arraystretch}{1.25}
\begin{tabular}{c|cccc}
\hline
Fine-tuning steps & PEACH & MLM & MLM with Reordering & Sample count \\
\hline
0	& 12.895563 & - & 7.420614 & 50K \\
1K & 13.166867 & -  & 8.080425 & 50K  \\
3K	& 15.16206 & -  & 14.288106 & 50K  \\
5K	& 20.924359 & -  & 17.13509 & 50K  \\
10K	& 26.17157 & -  & 21.928285 & 50K  \\
15K	& 27.991698 & -  & 24.24655 & 50K  \\
20K	& 28.940293 & -  & 25.608678 & 50K  \\
25K	& 29.325823 & -  & 26.304938 & 50K  \\
30K	& 29.727194 & -  & 26.939679 & 50K  \\
35K	& 30.017766 & -  & 27.462619 & 50K  \\
40K	& 30.122644 & -  & 27.770637 & 50K  \\
45K	& 30.228142 & -  & 28.110951 & 50K  \\
50K	& 30.491127 & -  & 28.309444 & 50K  \\
\hline
50K & 31.251482 & 21.384103 & 29.029701 & 5M \\
\hline
\end{tabular}
\caption{The zero- and few-shot translation results of the models evaluated with BLEU score on EN-FR section of WMT 14. These numbers are reported in Figure \ref{fig:zero-few-experiment-steps}}
\label{appendix:table:few-results}
\end{table*}

\end{document}